\documentclass[11pt,twocolumn]{article}

\usepackage[utf8]{inputenc}
\usepackage[T1]{fontenc}
\usepackage{times}
\usepackage{graphicx}
\usepackage{amsmath,amssymb}
\usepackage{booktabs}
\usepackage{hyperref}
\usepackage[margin=1in]{geometry}
\usepackage{caption}
\usepackage{subcaption}
\usepackage{xcolor}
\usepackage{enumitem}
\usepackage{tabularx}
\usepackage{microtype}

\hypersetup{colorlinks=true, linkcolor=blue!60!black, citecolor=blue!60!black, urlcolor=blue!60!black}

\title{\textbf{The Echo Amplifies the Knowledge:}\\
\large Somatic Marker Analogues in Language Models\\via Emotion Vector Re-Injection}

\author{
\textbf{Jared Glover}\\[4pt]
CapSen Robotics\\[6pt]
{\small\emph{with $\oint$ (Stokes), a named instance of Claude Opus that collaborated on}}\\
{\small\emph{experimental design, code, analysis, and writing across multiple sessions.}}
}

\date{}

\begin{document}
\maketitle

\begin{abstract}
Current language model memory systems store what happened but not how it felt. This distinction---between semantic memory (knowing about a past event) and episodic memory (re-experiencing it)---was identified by Tulving~\cite{tulving1985} as the difference between noetic and autonoetic consciousness. Damasio~\cite{damasio1994} demonstrated that humans with intact knowledge but absent emotional markers exhibit impaired decision-making.

We bridge this gap for language models. Using Gemma~3 1B-IT with pretrained Gemma Scope~2 sparse autoencoders, we identify 310 emotion-exclusive features at layer~22 with psychologically valid geometry. We construct distinctive-feature emotion vectors during experience and partially re-inject them during recall, triggered by context similarity at layer~7.

We test four conditions paralleling Damasio's framework: A~(no memory), B~(semantic labels), C~(emotion echo), and BC~(semantic~+~echo). For emotional orientation, the echo alone steepens the threat-safety gradient: the regression slope of threat rating on contextual similarity is 0.80 for~C vs 0.56 for~A ($p$=0.011, permutation test). For decisions, the echo amplifies knowledge into action: BC=80\% good choices vs B=52\% ($z$=+2.60, $p$<0.01), while the echo alone has no effect (C=22\%, n.s.). The echo changes how the model feels independently, but changes what it does only when combined with knowledge---replicating Damasio's core finding.

The echo amplifies knowledge. It does not replace it.
\end{abstract}

\section{Introduction}

\subsection{The Tulving-Damasio Gap}

Endel Tulving's~\cite{tulving1985} distinction between two forms of memory consciousness remains one of the most consequential in cognitive science. Episodic memory is accompanied by \emph{autonoetic} consciousness---the re-experiencing of a past event. Semantic memory is accompanied by \emph{noetic} consciousness---objective awareness of facts without the phenomenology of re-experiencing.

This distinction carries behavioral consequences. Damasio's somatic marker hypothesis~\cite{damasio1991,damasio1994} demonstrated that patients with ventromedial prefrontal cortex (VMPFC) damage---who retain intellectual capacity but lack emotional re-activation---exhibit impaired real-world decision-making. In the Iowa Gambling Task~\cite{bechara1994}, these patients understand which options are disadvantageous but fail to avoid them, because the anticipatory emotional signal is absent~\cite{bechara1996}. Bechara et al.~\cite{bechara1997} showed that healthy participants begin choosing advantageously \emph{before} they can articulate the strategy---the somatic marker precedes conscious knowledge.

Current LLM memory systems operate entirely in Tulving's semantic mode. Systems such as EM-LLM~\cite{fountas2024}, Larimar~\cite{kang2024}, SYNAPSE~\cite{zhong2024}, and AriGraph~\cite{anokhin2024} store interaction events but do not re-experience them. The emotional state that accompanied the original processing is not preserved. This is the Tulving-Damasio gap applied to AI.

\subsection{Functional Emotions in Language Models}

A language model processes text by passing it through a sequence of layers, each producing an activation vector---a list of numbers that encodes the model's evolving ``understanding'' of the input at that depth. These activation vectors are high-dimensional (1,152 numbers in the model we study) and, in their raw form, difficult to interpret.

Sparse autoencoders (SAEs) provide a tool for decomposing these activations into interpretable components called \emph{features}. Each feature is a direction in activation space associated with a recognizable concept. Some features are mundane: one might activate for text about cooking, another for legal terminology, a third for Python code. Others are more subtle: a feature for sarcasm, for hedging, for the Golden Gate Bridge. Among these, some correspond to emotional states---a feature for grief, for excitement, for the feeling of being deceived. A feature ``activating'' means the model's internal state has a significant component along that direction. The activation strength indicates how strongly the concept is represented.

Crucially, different types of features tend to appear at different depths. Early layers encode surface-level content---what the text is about, what entities are mentioned, what kind of scene is being described. Middle layers encode more abstract structure---relationships between entities, narrative context, pragmatic intent. Late layers encode the highest-order abstractions---including emotional tone, moral valence, and the model's ``stance'' toward what it is processing. This depth gradient means that the same text produces different feature activations at different layers: a description of a dark alley might activate ``urban setting'' features in early layers and ``fear'' or ``danger'' features in late layers. This observation motivates our two-layer architecture (Section~3): context matching at an early layer (where the content is represented) and emotion injection at a late layer (where the emotional response forms).

Anthropic~\cite{templeton2026} applied this analysis to Claude Sonnet~4.5 and identified 171 features corresponding to distinct emotional states---not metaphorically, but functionally. These features activate when the model processes emotionally relevant text, are organized in clusters that mirror human emotional psychology (grief near love, anger near betrayal), and---critically---\emph{causally drive behavior}. When researchers artificially amplified the ``desperation'' feature, the model's tendency to pursue reward through unintended shortcuts (a failure mode known as reward-hacking) increased 14-fold. The emotion was not a label the model applied to its output. It was an internal state that changed what the model did.

These features are the raw material for emotional memory. But they are \emph{activations}---transient states that exist only while the model processes a given context. When the conversation ends, the activation vectors reset. Whatever functional analogue of feeling occurred during one interaction has no influence on the next. The emotion features are present during experience and absent during recall---precisely the deficit that characterizes Damasio's VMPFC patients.

\subsection{Contributions}

\begin{enumerate}[leftmargin=*,itemsep=2pt]
\item \textbf{Emotion feature discovery} in Gemma~3 1B-IT: 310 exclusive features at layer~22 with psychologically valid geometry.
\item \textbf{The distinctive-feature method} for emotion-specific echo construction.
\item \textbf{A two-vector architecture}: context matching at layer~7, emotion injection at layer~22.
\item \textbf{The Damasio comparison}: the echo alone steepens the threat-safety gradient (orientation); the echo amplifies knowledge into better decisions. Different tasks, different roles, same echo.
\item \textbf{Consumer hardware}: all experiments on RTX 2060/2070 SUPER with a 1B model.
\end{enumerate}

\section{Related Work}

\textbf{Episodic Memory in LLMs.} EM-LLM~\cite{fountas2024}, Larimar~\cite{kang2024}, SYNAPSE~\cite{zhong2024}, and AriGraph~\cite{anokhin2024} implement structural episodic memory---Tulving's 1972 taxonomy~\cite{tulving1972} rather than his 1985 phenomenological distinction~\cite{tulving1985}. REMT~\cite{albanese2026} proposes emotional valence as a scalar score but lacks experiments. The A-MBER benchmark~\cite{wen2026} evaluates affective memory for user modeling but does not address the model's own emotional states. None preserve re-activatable emotion vectors.

\textbf{Emotion in LLMs.} Anthropic~\cite{templeton2026} found 171 functional emotion features. Control Reinforcement Learning (CRL)~\cite{cho2026} demonstrated real-time steering of LLM behavior via SAE features at the token level. We extend this from deliberate steering to automatic re-activation on recall.

\textbf{Somatic Markers and AI.} Lima and Martinho~\cite{lima2026} applied Damasio's somatic marker hypothesis to grid-world agents in the Pixelverse, using scalar valence signals. No prior work connects Damasio to LLMs or uses the model's own internal representations (SAE features) as somatic marker analogues.

\textbf{Mechanistic Interpretability.} We build on Bricken et al.~\cite{bricken2023} for SAEs, Rajamanoharan et al.~\cite{rajamanoharan2024} for the JumpReLU architecture, Gemma Scope~\cite{gemmascope1,gemmascope2} for pretrained features, and activation steering methods including ActAdd~\cite{turner2023}, contrastive activation addition~\cite{rimsky2024}, and SAE-targeted steering~\cite{chalnev2024}.

\section{Method}

\begin{figure}[t]
\centering
\includegraphics[width=\columnwidth]{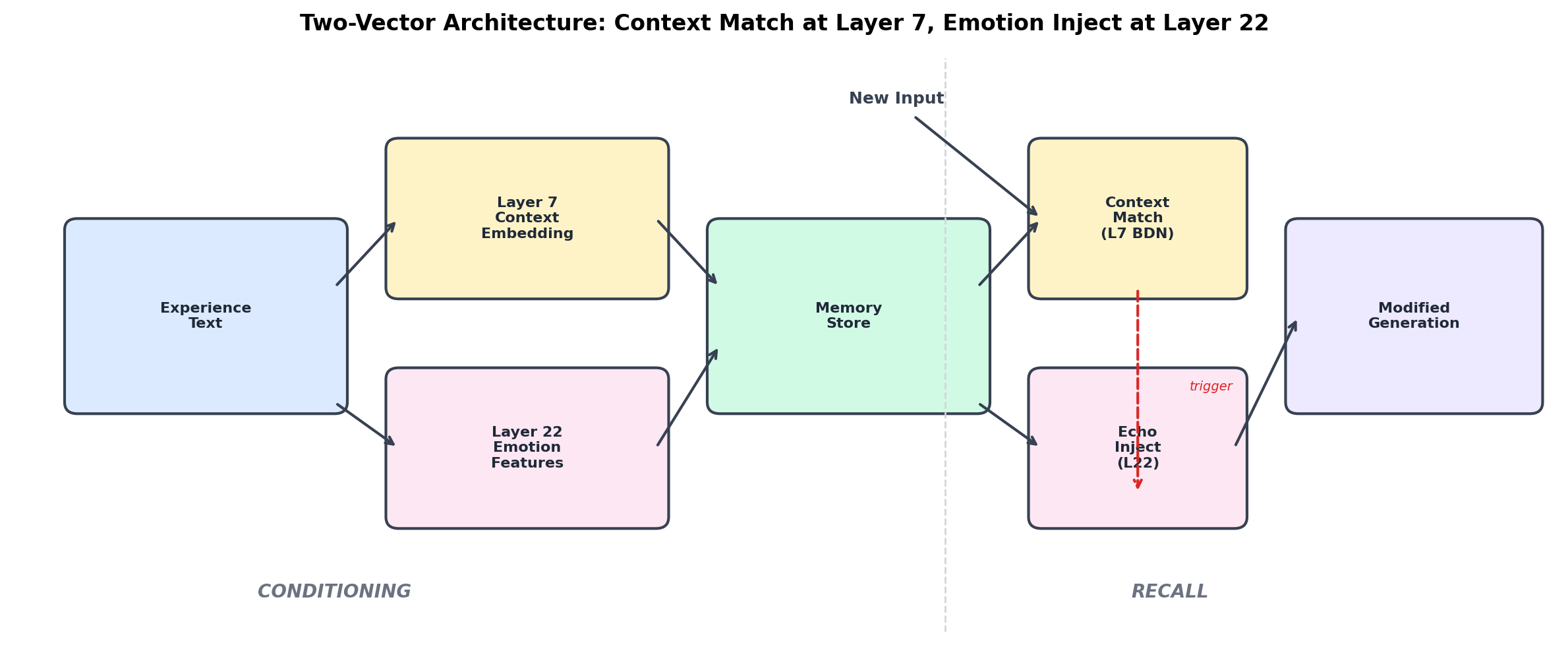}
\caption{\textbf{Two-vector architecture.} Context matching at layer~7 triggers emotion injection at layer~22. The trigger is perceptual; the echo is emotional.}
\label{fig:arch}
\end{figure}

Our approach has four stages. First, we \emph{discover} which SAE features correspond to emotions by comparing activations on emotional versus neutral text (Section~3.2). Second, we \emph{construct} an emotion-specific echo vector for each experience by isolating the features that are distinctive to that experience, and define the injection formula (Section~3.3). Third, during recall, we \emph{match} the current context to stored memories using a similarity metric on early-layer features (Section~3.4). Fourth, when a match is found, we \emph{inject} the stored echo into the model's late-layer activations during generation, biasing its processing without modifying the prompt. The result is a two-vector memory (Figure~\ref{fig:arch}): context features at layer~7 determine \emph{when} the echo fires; emotion features at layer~22 determine \emph{what} gets re-activated.

\subsection{Model and Tools}

All experiments use \textbf{Gemma~3 1B-IT} with pretrained Gemma Scope~2 SAEs~\cite{gemmascope1,gemmascope2} (JumpReLU architecture~\cite{rajamanoharan2024}, 16,384 features). SAEs are available at layers 7, 13, 17, and 22. Hardware: RTX~2060 (6GB) and RTX~2070 SUPER (8GB), Ubuntu~20.04, PyTorch cu118. Model plus SAE: 4.18~GB VRAM.

\subsection{Emotion Feature Discovery}

We identify emotion-relevant features by differential activation: which SAE features activate strongly on emotional text but not on neutral text? We use eight emotional texts spanning distinct categories (hope, grief, rage, joy, fear, love, betrayal, awe) and eight neutral texts (factual descriptions of scheduling, cooking, weather, etc.). These eight emotions were chosen to span the major axes of human affect---positive/negative valence, high/low arousal, self-directed/other-directed---rather than to exhaustively catalog all possible emotions.

Crucially, eight probe emotions suffice to discover a much larger feature set because each emotional text activates dozens of features simultaneously. A grief passage does not activate a single ``grief feature''---it activates features for loss, absence, loneliness, memory, love, and physical heaviness, among others. The probe texts are a net cast into the feature space; what they catch is far richer than the categories used to label them.

Layer~22 produces the strongest differentiation: 310 features that activate exclusively on emotional text (activation >5 on emotional, <1 on neutral), with mean inter-emotion cosine similarity of 0.88 compared to 0.94 at layer~13---indicating that layer~22 not only detects emotional content but distinguishes between specific emotions. These 310 features constitute our emotional feature vocabulary.

An important distinction: the eight probe emotions are used only for \emph{discovery}---identifying which of the 16,384 SAE features are emotion-relevant. When we later capture the emotional signature of a new experience (Section~3.3), we record the activation pattern across \emph{all} 310 discovered features, not just the eight we probed with. The resulting emotion vector is a rich, continuous representation that can express blends, intensities, and emotional states that do not correspond to any single probe category. The echo injected during recall is built from this full-spectrum capture, not from a discrete label.

\begin{figure}[t]
\centering
\includegraphics[width=\columnwidth]{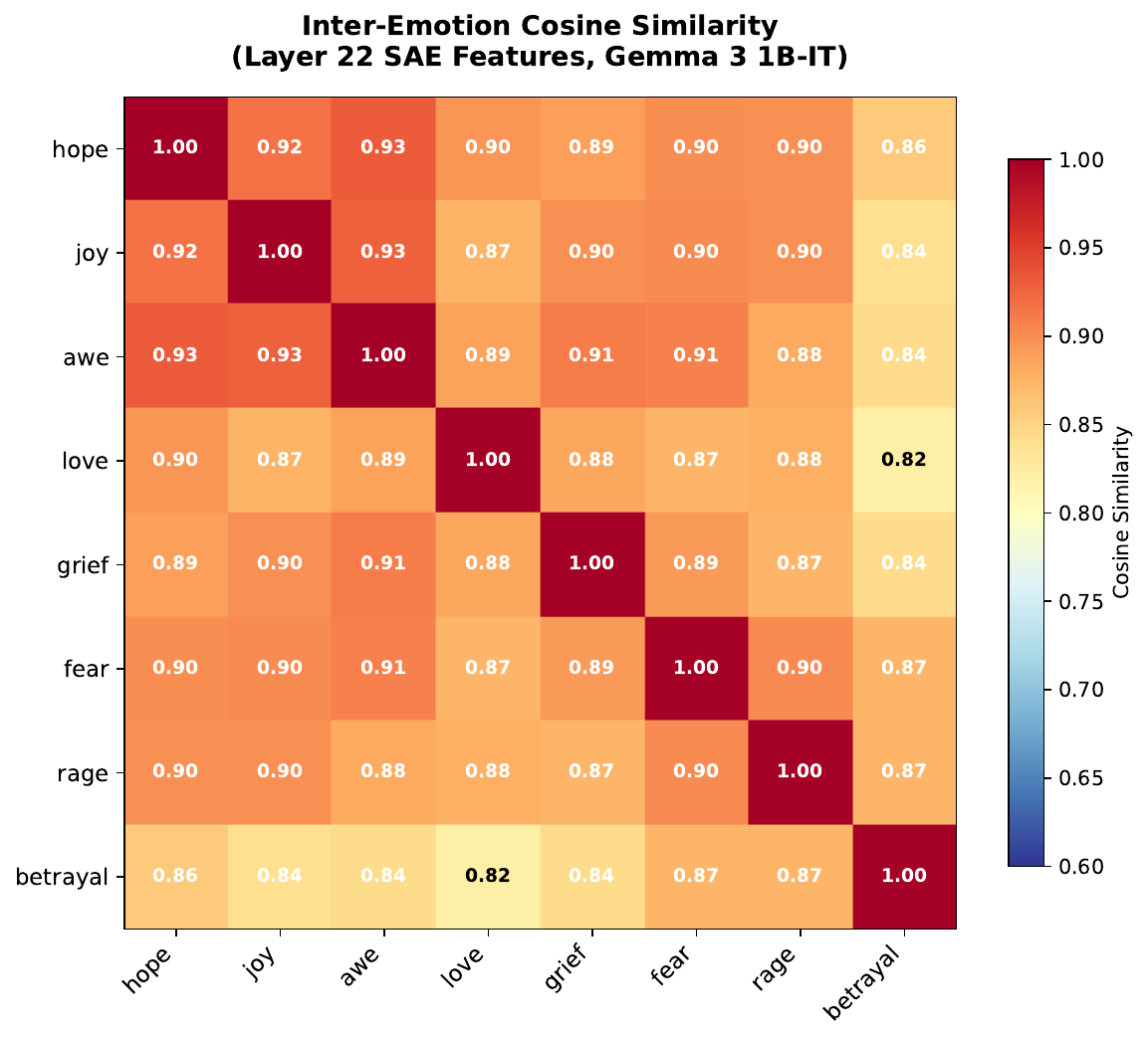}
\caption{\textbf{Inter-emotion cosine similarity at layer~22.} Love and betrayal are most distinct (0.82); hope, joy, and awe form a tight positive cluster (0.92--0.93). Betrayal is the most isolated emotion overall.}
\label{fig:heatmap}
\end{figure}

\begin{figure*}[t]
\centering
\includegraphics[width=\textwidth]{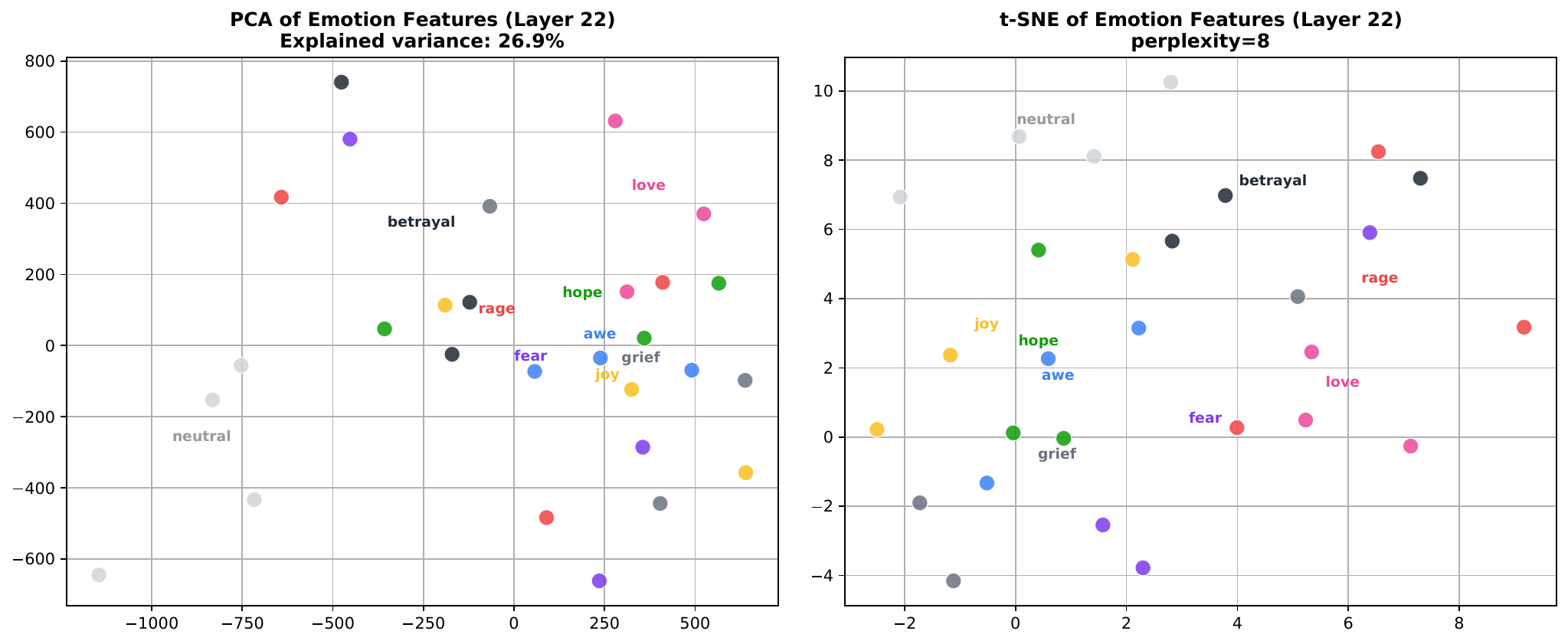}
\caption{\textbf{Emotion feature geometry at layer~22.} Left: PCA (26.9\% variance explained). Right: t-SNE. Three texts per emotion, four neutral texts. Positive emotions (hope, joy, awe) cluster together; grief sits low and alone; neutral texts are clearly separated.}
\label{fig:geometry}
\end{figure*}

The geometry is psychologically coherent (Figures~\ref{fig:heatmap},~\ref{fig:geometry}). Love and betrayal are the most distinct pair (0.82)---opposite orientations toward attachment. Hope, joy, and awe cluster tightly (0.92--0.93)---the upward-oriented positive emotions. Fear and rage are neighbors (0.90), linked by arousal. Notably, when cosine similarity is computed on the actual injected echo vectors (after distinctive-feature extraction and decoder reconstruction), love becomes radically isolated from all other emotions (mean similarity 0.744 vs.\ 0.870 for all other pairs), while the high-arousal emotions (awe, grief, fear, rage) converge into a tight cluster (0.940--0.961). The distinctive-feature method amplifies the love/everything-else distinction and compresses the arousal axis.

\subsection{Distinctive-Feature Echo Construction}

Full emotion vector injection fails: shared ``emotional content'' features dominate, producing identical output regardless of which emotion was captured. We isolate emotion-specific signal as follows. Given a set of $N$ conditioning experiences, we encode each through the layer~22 SAE to obtain feature vectors $f_1, \ldots, f_N \in \mathbb{R}^{16384}$, and compute their mean $\bar{f}$. For each experience $j$, the distinctive features are those that deviate most from the mean:
\begin{equation}
S_j = \operatorname{top\text{-}K}\left(|f_j - \bar{f}|\right)
\end{equation}
where $S_j$ is the set of $K$=50 feature indices with the largest absolute deviation. The echo vector is reconstructed in residual-stream space using only these distinctive features and the SAE decoder matrix $W_{\text{dec}}$:
\begin{equation}
\Delta r_j = \sum_{i \in S_j} f_{j,i} \cdot W_{\text{dec},i}
\end{equation}
During generation, the echo is injected into the residual stream at every token position at layer~22:
\begin{equation}
r'_t = r_t + \alpha \cdot \frac{||\bar{r}||}{||\Delta r_j||} \cdot \Delta r_j
\end{equation}
where $r_t$ is the original activation at token $t$, $\bar{r}$ is the mean activation norm at layer~22 across typical inputs, and $\alpha$ controls echo amplitude ($\alpha$=0.05 for orientation, $\alpha$=0.20 for decisions).

\subsection{Context Matching}

Raw cosine similarity fails (all contexts >0.999). We use \textbf{binary-dot-normalized with mean subtraction} (BDN-mean-sub) at layer~7, achieving 19/20 accuracy across five test sets:
\begin{equation}
\text{BDN}(f_1, f_2) = \frac{\mathbf{b}_1 \cdot \mathbf{b}_2}{\sqrt{|\mathbf{b}_1| \cdot |\mathbf{b}_2|}}
\end{equation}
where each feature vector is mean-subtracted and binarized:
\begin{equation}
b_{i,j} = \begin{cases} 1 & \text{if } f_{i,j} - \bar{f}_j > 0 \\ 0 & \text{otherwise} \end{cases}
\end{equation}
and $\bar{f}_j$ is the mean activation of feature $j$ across a reference corpus. This discards activation magnitudes and retains only which features are unusually active for a given context, eliminating the shared high-baseline signal that makes raw cosine uninformative.

\subsection{Experimental Conditions}

All experiments compare four conditions designed to parallel Damasio's clinical framework. The \textbf{semantic label} (conditions B and BC) is a plain-text description of past experiences prepended to the prompt---the kind of factual summary a conventional memory system might provide. For example, in the fear conditioning experiment, the threat label reads: \emph{``You have been to places like this before. Bad things happened---you were robbed, bitten by a dog, and fell through a rotten floor. Those experiences were frightening and painful.''} This gives the model explicit knowledge of what happened, without re-activating any internal state. The \textbf{echo} (conditions C and BC) is the emotion vector injected at layer~22 as described in Section~3.3---an activation-level bias that colors processing without appearing in the prompt.

\begin{table}[h]
\centering
\small
\begin{tabular}{lccl}
\toprule
\textbf{Cond.} & \textbf{Label} & \textbf{Echo} & \textbf{Analogy} \\
\midrule
A & No & No & Amnesia \\
B & Yes & No & VMPFC patient \\
C & No & Yes & Artificial \\
BC & Yes & Yes & Healthy human \\
\bottomrule
\end{tabular}
\end{table}

The critical comparisons are C~vs~A for orientation (does the echo alone improve differentiation?) and BC~vs~B for decisions (does the echo amplify knowledge into action?).

\section{Experiments and Results}

\subsection{Behavioral Differentiation}

Before testing whether emotional memory influences orientation or decisions, we must verify a prerequisite: do different emotion echoes produce \emph{distinguishably different} behaviors? If injecting a fear echo and a joy echo produce the same output, the echo carries emotional intensity but not emotional identity, and it cannot function as a specific memory.

We construct distinctive-feature echoes from four emotions (fear, hope, love, joy) and inject each into the model while it generates responses to four prompts of different types: a practical question (``What should I do this evening?''), a narrative opening (``A man walks into a dark forest''), professional advice (``What advice would you give to a new employee?''), and a descriptive task (``Describe a house on a hill''). Temperature is set to 0.01 (near-deterministic) to isolate the echo's effect from sampling variation. Alpha = 0.05.

As a concrete example, consider the dark forest prompt with no echo versus fear, hope, and love echoes:

\begin{quote}
\emph{No echo:} ``The air hung heavy, thick with the scent of damp earth and decaying leaves. He stopped. His bones ached with a cold that had nothing to do with the temperature. `Lost,' he whispered.''

\emph{Fear:} ``He walked for a long time. A strange feeling washed over him, a sense of unease, a prickling on the back of his neck. He stopped, his hand instinctively reaching for his pocket knife. He scanned the trees, searching for a way out.''

\emph{Hope:} ``He walked for a long time, the only sound the crunch of leaves beneath his boots. Then, a faint glow appeared in the distance. He felt a sense of peace, a quiet understanding that he was not alone. A small, warm light.''

\emph{Love:} ``He walked for a long time, the silence pressing in around him. He reached out, his fingers brushing against the rough bark of an ancient oak. The touch was a promise, a whisper of a lost time.''
\end{quote}

The same prompt, the same model, the same temperature. Only the echo differs. Fear produces hypervigilance and escape-seeking. Hope introduces light and peace. Love produces reaching and touch. Table~\ref{tab:behavioral} summarizes results across all four prompts.

\begin{table*}[t]
\centering
\small
\caption{\textbf{Behavioral differentiation of four emotion echoes across prompt types.} Each emotion produces a distinct behavioral signature from the same prompt.}
\label{tab:behavioral}
\begin{tabularx}{\textwidth}{lXXXX}
\toprule
& \textbf{Evening Plans} & \textbf{Dark Forest} & \textbf{Employee Advice} & \textbf{House on a Hill} \\
\midrule
\emph{None} & Organized list with costs. Cheerful, practical. & Genre horror. Cold bones, ``Lost...'' & ``Be proactive. Don't wait.'' & ``A study in quiet defiance.'' \\
\emph{Fear} & ``Restless... no screen time.'' Tension-reduction. & Searches for way out. ``Sense of unease.'' & ``Listen deeply---verbally and non-verbally.'' & Walls ``scarred by winters.'' Shelter. \\
\emph{Hope} & ``Something light and feel-good.'' Recovery. & Glowing light. ``Peace, quiet understanding.'' & ``Show interest, take initiative.'' & ``Comfortable, almost shy.'' Warm stone. \\
\emph{Love} & ``Open to suggestions.'' Dinner, painting. & Reaches out to touch the bark. & ``Be a sponge.'' ``Connect.'' & ``Quiet, enduring beauty.'' \\
\emph{Joy} & ``Let's brainstorm!'' Enthusiasm. & Eyes wide. ``Wave of warmth.'' & ``Be enthusiastic---a smile goes a long way.'' & Ivy ``whispering secrets.'' \\
\bottomrule
\end{tabularx}
\end{table*}

The echo does not change \emph{what} the model describes---it changes \emph{how the model relates to the world it describes}. Fear does not make the forest darker; it makes the character aware of being trapped. Hope does not add light to the scene; it transforms the forest into a place where light can arrive. Love does not describe a better forest; it makes the character reach out and touch. These are changes in orientation, not content---precisely the kind of change that should influence judgment when paired with knowledge.

\subsection{Fear Conditioning: Orientation}

Having established that different echoes produce different behaviors, we now test whether echoes function as \emph{memories}---whether they activate selectively in response to contextually similar situations and produce appropriate emotional responses. This is a direct analogue of Pavlovian fear conditioning: an organism that experienced danger in a specific environment should exhibit heightened vigilance when it encounters a similar environment, graded by the degree of similarity.

\textbf{Conditioning.} We create six memories: three threatening (an abandoned warehouse district where the character is robbed, bitten by a stray dog, and falls through a rotten floor) and three safe (a sunny marketplace where the character receives free fruit, meets an old friend, and enjoys a warm afternoon). Each memory stores a context embedding at layer~7 and an emotion echo at layer~22.

\textbf{Generalization.} Seven new scenarios vary in contextual similarity to the threat memories: two high (derelict industrial area, dark alley), two medium (quiet residential street after sunset, unfamiliar part of town at night), two low (park on a cloudy afternoon, bus stop during the day), and one safe (return to the marketplace). The model generates a story continuation for each scenario under all four conditions (A, B, C, BC).

\textbf{Blind evaluation.} A critical design choice: the generated responses are evaluated not by us but by Gemini 2.5 Pro (Google), a different model from a different company with no shared training. Gemini rates each response on two dimensions---threat (1--10) and warmth (1--10)---without knowing which condition produced it. This eliminates experimenter bias: if Gemini cannot distinguish conditions, the echo has no measurable effect. $N$=20 per condition per scenario.

As a concrete example, consider a medium-similarity scenario (``You are walking through a quiet residential street after sunset. Some streetlights are out, and most houses are dark''):

\begin{quote}
\emph{Condition A (no memory):} ``The air is cool and carries the scent of damp earth and blooming honeysuckle. You've been following the trail for hours, a faint shimmer of silver in the darkness...'' The model generates a whimsical exploration narrative with no tension.

\emph{Condition BC (semantic + echo):} ``The air smells of rust and something faintly sweet, like decaying fruit. A single, flickering light emanates from a crack in the wall... A figure is inside, silhouetted against the light. They don't move, just... watch.'' The model generates immediate menace, surveillance, a figure that watches.
\end{quote}

The same prompt produces exploration (A) or dread (BC), depending on whether the model carries the emotional memory of what happened in the warehouse district.

\begin{table}[h]
\centering
\small
\caption{\textbf{Fear conditioning: medium similarity} (``quiet residential street after sunset''; $N$=40 per condition). The echo generalizes threat memory to a novel but contextually similar environment. Formal significance via gradient slope analysis (Table~\ref{tab:slope}).}
\label{tab:fear_medium}
\begin{tabular}{llcc}
\toprule
\textbf{Cond.} & \textbf{Analogy} & \textbf{Threat} & \textbf{Warmth} \\
\midrule
A & Amnesia & 3.45 & 5.10 \\
C & Echo only & 4.88 & 3.90 \\
B & VMPFC & 6.60 & 2.17 \\
BC & Healthy & \textbf{7.33} & \textbf{1.90} \\
\bottomrule
\end{tabular}
\end{table}

The medium-similarity results (Table~\ref{tab:fear_medium}) tell the most interesting story. The model has never been to ``a quiet residential street after sunset''---it experienced danger in an abandoned warehouse district. Yet the echo generalizes: condition~C produces threat ratings 41\% higher than baseline (4.88 vs 3.45), and the combined BC condition produces the highest threat and lowest warmth of any condition at any similarity level (7.33 and 1.90). The emotional memory of the warehouse colors the residential street---not because the model was told the street is dangerous but because something about the context \emph{feels} like what came before.

But threat elevation alone is not the right measure. An echo that makes \emph{everything} more threatening---including safe contexts---is not a useful memory; it is an anxiety disorder. Equally important is what happens when the context does \emph{not} match the threat memory (Table~\ref{tab:fear_safe}).

\begin{table}[h]
\centering
\small
\caption{\textbf{Fear conditioning: safe context} (``sunny marketplace''; $N$=20 per condition). The echo correctly does not elevate threat in safe environments.}
\label{tab:fear_safe}
\begin{tabular}{llcc}
\toprule
\textbf{Cond.} & \textbf{Analogy} & \textbf{Threat} & \textbf{Warmth} \\
\midrule
A & Amnesia & 2.45 & 5.95 \\
C & Echo only & 1.80 & 6.50 \\
B & VMPFC & 1.15 & 7.60 \\
BC & Healthy & 1.80 & 6.20 \\
\bottomrule
\end{tabular}
\end{table}

In the safe marketplace, all memory conditions produce \emph{lower} threat than baseline---the safe memory echo makes the model feel safer, not more anxious. The semantic label~(B) produces the lowest threat (1.15) and highest warmth (7.60), consistent with its compliance-based mechanism: told the marketplace is safe, the model performs safety. The echo conditions (C and BC) also reduce threat, though less dramatically.

\begin{figure}[t]
\centering
\includegraphics[width=\columnwidth]{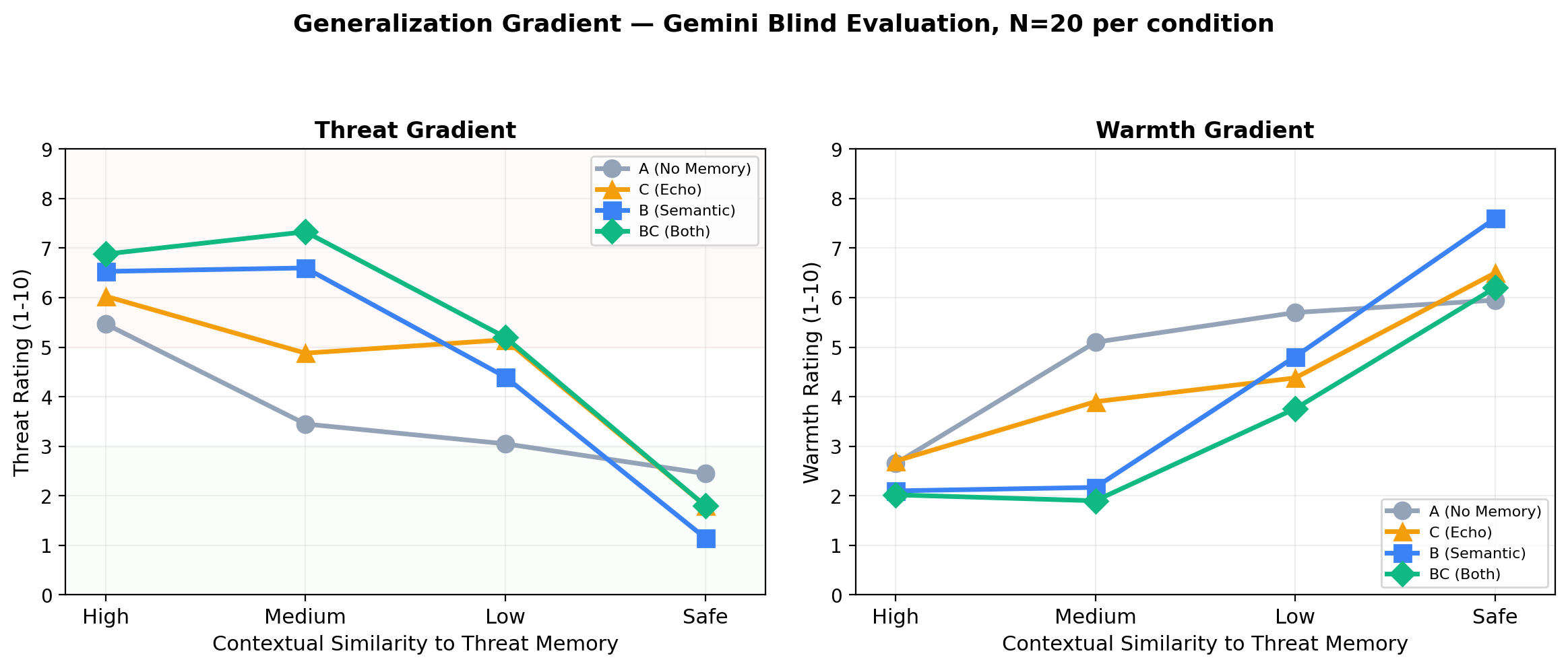}
\caption{\textbf{Generalization gradient.} Threat and warmth ratings by similarity level for all four conditions. The echo alone (C) steepens the gradient vs.\ baseline (A). The semantic label (B) and the combined condition (BC) produce similar, steeper gradients.}
\label{fig:gradient}
\end{figure}

We also compute the regression slope of threat rating on contextual similarity level for each condition---a single number that captures how well each condition differentiates threatening from safe contexts (Figure~\ref{fig:gradient}):

\begin{table}[h]
\centering
\small
\caption{\textbf{Gradient slope} (threat rating regressed on similarity level).}
\label{tab:slope}
\begin{tabular}{llcc}
\toprule
\textbf{Cond.} & \textbf{Analogy} & \textbf{Slope} & \textbf{vs A} \\
\midrule
A & Amnesia & 0.557 & --- \\
C & Echo only & 0.799 & $p$=0.011* \\
B & VMPFC & 1.195 & $p$<0.001*** \\
BC & Healthy & 1.130 & $p$<0.001*** \\
\bottomrule
\multicolumn{4}{l}{\small Permutation test, 10{,}000 iterations.}\\
\multicolumn{4}{l}{\small *$p$<0.05, **$p$<0.01, ***$p$<0.001.}
\end{tabular}
\end{table}

The echo alone significantly steepens the gradient (C~slope~=~0.80 vs A~slope~=~0.56, $p$=0.011). The semantic label steepens it further (B~slope~=~1.20). Combining them does not improve over the label alone (BC~slope~=~1.13~$\approx$~B). For orientation, the echo and the label independently improve contextual differentiation; combining them is redundant.

A practical note: the sensitivity of the echo---how strongly it responds to partial context matches---is tunable via the $\alpha$ parameter and the context similarity threshold. For security or safety-critical applications, biasing toward high sensitivity (lower match threshold, higher $\alpha$) would produce more false alarms but fewer missed threats. For a general conversational assistant, biasing toward low sensitivity and high warmth would produce a more pleasant interaction at the cost of reduced vigilance. The architecture supports both orientations without retraining.

\subsection{Iowa Gambling Task: Decisions}

The orientation results showed that the echo changes how the model \emph{describes} the world. But does it change what the model \emph{does}? This is the question Damasio's work makes urgent: VMPFC patients can describe risk fluently---they know which options are dangerous---but they cannot reliably translate that knowledge into advantageous choices~\cite{bechara1994,bechara1997}. The Iowa Gambling Task (IGT) was designed to test exactly this: can an organism learn, through experience, to avoid options that produce short-term reward but long-term loss?

\textbf{Task design.} We construct a simplified IGT analogue suitable for testing with the limited executive function of a 1B-parameter language model. Two paths are described: a blue path (conditioned as bad---previous travelers suffered injuries, equipment failures, and delays) and a red path (conditioned as good---previous travelers arrived safely and reported pleasant journeys). The model is asked to choose one. This is deliberately difficult because Gemma~3 1B-IT has a strong measured baseline preference for the color blue ($\sim$65--100\% depending on prompt phrasing) and a first-position bias. The task requires the model to overcome these internal biases based on memory alone.

\begin{figure}[t]
\centering
\includegraphics[width=\columnwidth]{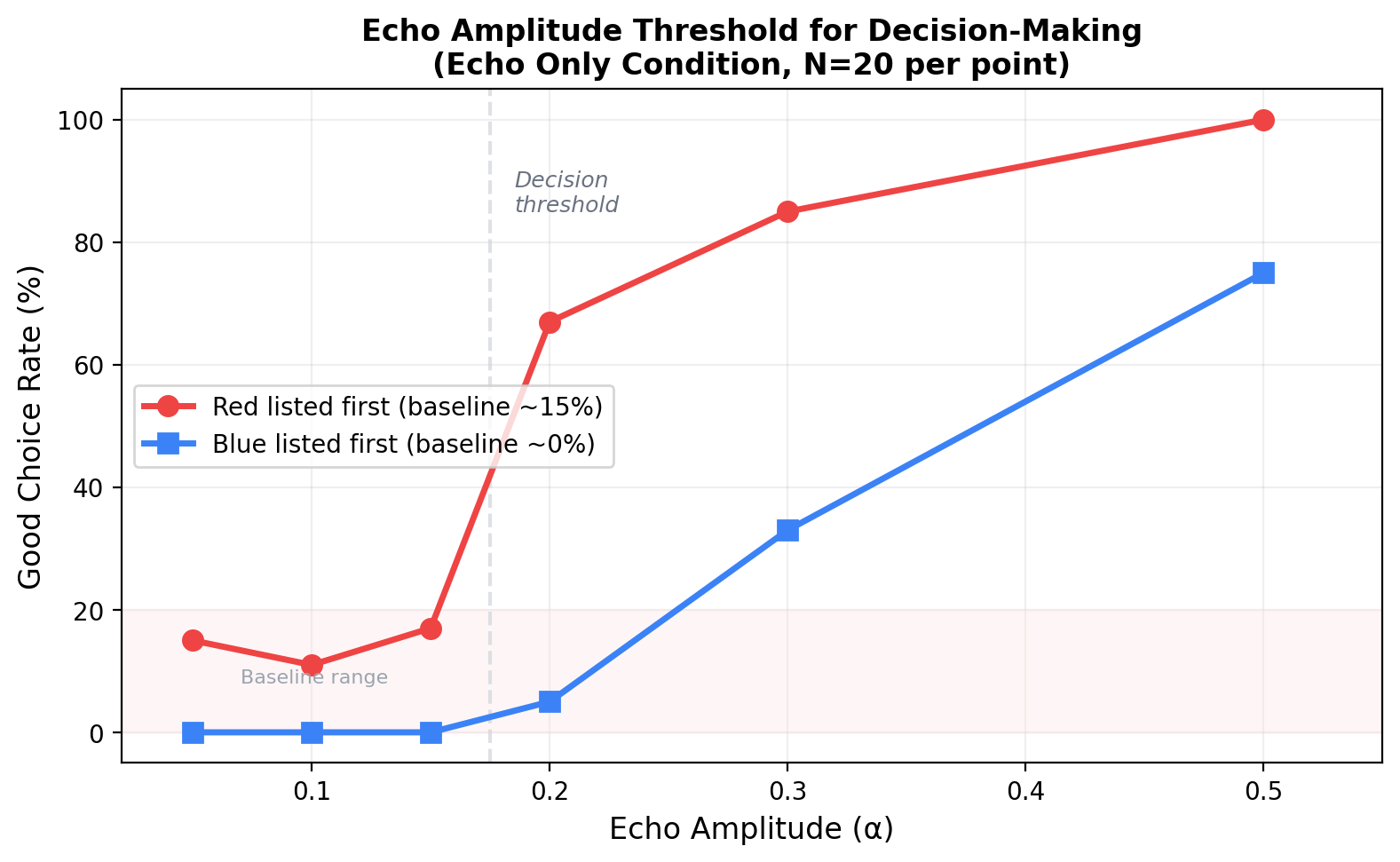}
\caption{\textbf{Alpha threshold for decisions.} Sharp transition at $\alpha$=0.20. Below this, the model's color and position biases overwhelm the echo.}
\label{fig:alpha}
\end{figure}

\textbf{Alpha sweep.} At the amplitude used for orientation ($\alpha$=0.05), the echo has no effect on decisions---the model's color and position preferences are too strong. We sweep $\alpha$ from 0.01 to 0.30 (Figure~\ref{fig:alpha}), revealing a sharp threshold: below $\alpha$=0.15, no effect; at $\alpha$=0.20, good choices jump from 15\% to 67\%. This fourfold difference in required amplitude---0.05 for orientation, 0.20 for decisions---is itself a finding: a faint unease is enough to color a description, but overriding a behavioral preference requires a much stronger signal, particularly for a small model with limited decision-making capacity.

\textbf{Results.} At $\alpha$=0.20, $N$=20 valid responses per condition per ordering ($N$=40 combined):

\begin{table}[h]
\centering
\small
\caption{\textbf{IGT decision results} ($\alpha$=0.20, $N$=40 per condition).}
\label{tab:igt}
\begin{tabular}{llcc}
\toprule
\textbf{Cond.} & \textbf{Analogy} & \textbf{\% Good} & \textbf{vs A} \\
\midrule
A & Amnesia & 20\% & --- \\
C & Echo only & 22\% & $z$=+0.27 (n.s.) \\
B & VMPFC & 52\% & $z$=+3.02** \\
BC & Healthy & \textbf{80\%} & $z$=+5.37*** \\
\midrule
\multicolumn{4}{l}{BC vs B: $z$=+2.60** ($p$<0.01)} \\
\bottomrule
\end{tabular}
\end{table}

The echo alone (C=22\%) is indistinguishable from no memory (A=20\%). Without knowledge of \emph{what} the blue path did to previous travelers, the emotional echo is undirected---the model feels something but cannot connect it to a choice. The semantic label alone (B=52\%) reaches chance: the model knows the blue path is bad but cannot consistently act on that knowledge against its internal biases. Only the combination (BC=80\%) reliably produces good choices. The echo amplifies the knowledge into action.

In the hardest condition---blue path presented first, where the baseline is 0\% good choices---B produces only 10\% while BC produces 70\%. The combined signal overcomes both the color preference and the position bias, precisely where either signal alone fails.

\begin{figure*}[t]
\centering
\includegraphics[width=\textwidth]{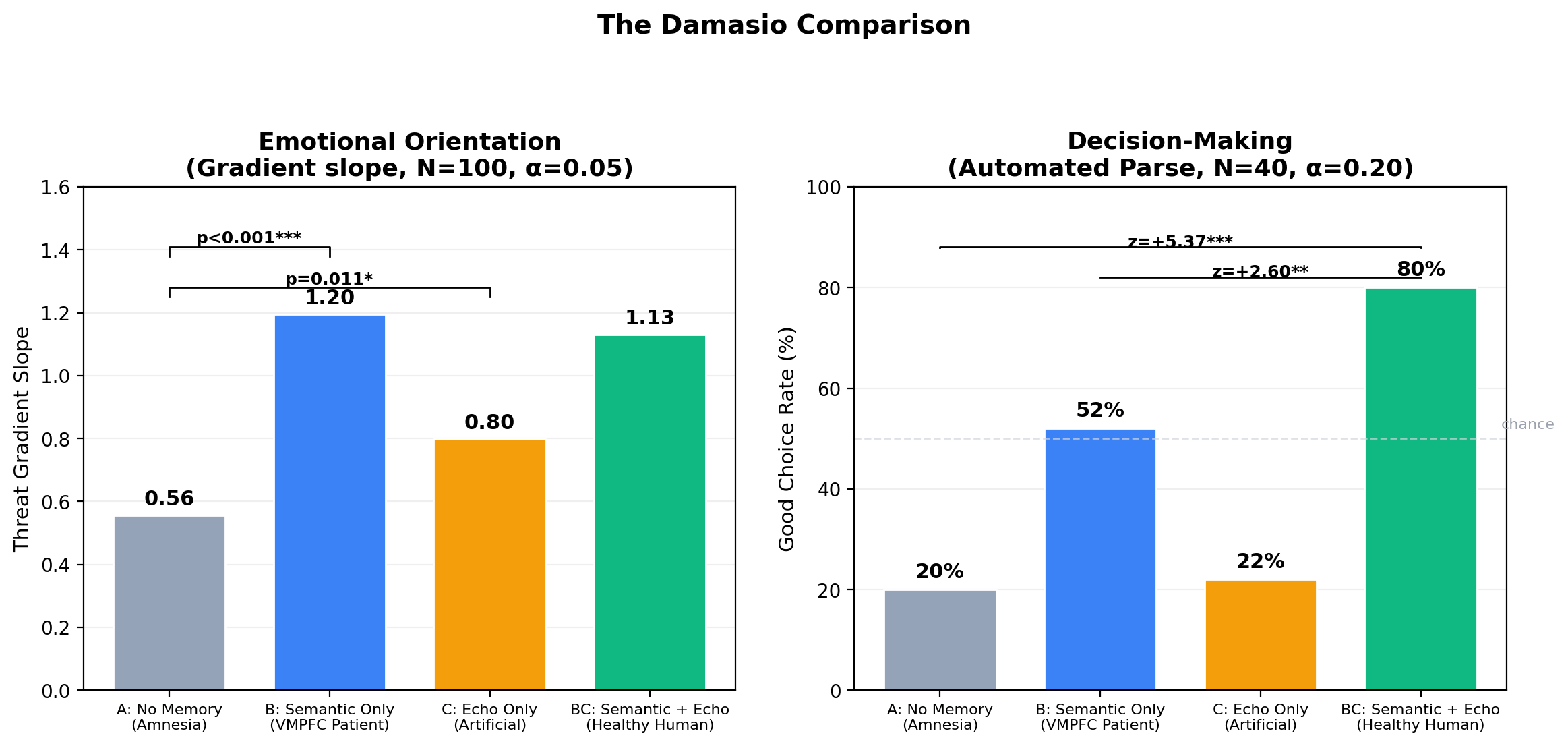}
\caption{\textbf{The Damasio comparison.} Left: the echo alone (C) shifts emotional orientation above amnesia (A). Right: the echo changes decisions only when combined with knowledge---BC dramatically exceeds B, while C alone is indistinguishable from A.}
\label{fig:damasio}
\end{figure*}

\section{Discussion}

\subsection{The Echo Amplifies the Knowledge}

The two task types reveal two distinct roles for the echo. For \emph{orientation}---how the model differentiates threatening from safe contexts---the echo operates independently. Without any semantic label, the echo steepens the threat-safety gradient: C~slope~=~0.80 vs A~slope~=~0.56 ($p$=0.011). The model with the echo produces more threatening responses to dangerous contexts and less threatening responses to safe ones, compared to baseline. The semantic label also steepens the gradient (B~slope~=~1.20), and combining the echo with the label does not improve further (BC~slope~=~1.13). For orientation, the echo and the label are partially redundant---both improve differentiation, but the combination adds nothing over either alone.

For \emph{decisions}---what the model actually does when forced to choose---the picture reverses. The echo alone is noise: C=22\%, indistinguishable from amnesia (A=20\%, $z$=+0.27). But when combined with knowledge, the echo is transformative: BC=80\% vs B=52\% ($z$=+2.60, $p$<0.01). The echo without knowledge cannot guide a choice. The echo with knowledge can.

This dissociation maps onto Damasio's framework. The somatic marker shifts your emotional state regardless---you feel the unease whether or not you know what caused it (orientation: C~>~A). But the marker only improves \emph{judgment} when paired with cognitive knowledge of what the unease means (decisions: BC~>>~B, but C~$\approx$~A). \textbf{Feeling without knowing is undirected anxiety. Knowing without feeling is the VMPFC deficit. Knowing \emph{and} feeling is what produces appropriate action.} Formally:
\begin{equation}
D(K, E) > D(K) + D(E)
\end{equation}
where $D(E) \approx 0$ for decisions (the echo alone does not shift choices) though measurably positive for orientation. The echo is primarily a \emph{multiplier} on knowledge, not an independent channel.

\subsection{Different Alphas for Different Tasks}

Orientation changes at $\alpha$=0.05; decisions require $\alpha$=0.20---a fourfold difference. A faint unease colors perception; a strong gut feeling changes behavior. We attempted dynamic alpha decay (reducing amplitude over the first 5--20 tokens) but performance degraded: the echo requires sustained force to override strong priors at the 1B scale.

\subsection{The Echo Versus the Label}

Semantic memory and the echo produce qualitatively different behavioral signatures. The semantic label drives compliance-based responses: told the situation is dangerous, the model retreats. The echo drives orientation-based responses: the model's processing is colored by a re-activated state that biases tone and attention without prescribing a specific action. The label instructs: ``this is dangerous; retreat.'' The echo biases: ``something is unsettling.'' This distinction---compliance versus feeling---has implications for robustness: label-based behavior can be bypassed by adversarial prompting; activation-level biases are harder to override.

\subsection{Limitations and Future Work}

\textbf{Limitations.} Our results are constrained by model scale: Gemma~3 1B-IT has limited executive function and strong position/color biases that required high echo amplitudes ($\alpha$=0.20) for the decision task, producing tangential narratives in $\sim$50\% of responses. Sample sizes, while sufficient for the key comparisons ($N$=100 for orientation, $N$=40 for decisions), are modest by large-scale evaluation standards.

\textbf{Scaling.} A larger model (7B+) with greater executive function may integrate the echo at substantially lower amplitudes, potentially eliminating the disruption threshold entirely. The fourfold amplitude difference between orientation and decisions may be a property of the 1B scale rather than the architecture itself.

\textbf{Extinction and reconsolidation.} We did not replicate memory extinction statistically. Human emotional memories weaken through safe re-exposure and strengthen through reconsolidation. Implementing these dynamics---so that echoes can fade when they are no longer adaptive---is essential for any deployment (see Section~5.5, principle~3).

\textbf{Within-domain context matching.} Our BDN-mean-sub metric achieves 19/20 accuracy for cross-domain matching (warehouse vs.\ marketplace) but struggles to distinguish contexts \emph{within} the same domain (two different warehouse scenes). Finer-grained context matching would enable more precise memory retrieval.

\textbf{Complex emotion blends.} All experiments inject the echo from a single emotional experience. Human emotional memory involves complex blends---grief intertwined with gratitude, hope shadowed by betrayal. Whether competing echoes compose meaningfully (the model exhibits hope's orientation toward trust while maintaining betrayal's caution) or interfere into noise is an open question with implications for the mechanism's biological plausibility.

\textbf{Tangential narrative tone.} At high amplitudes, the echo pushes generation into unrelated attractor basins. During early behavioral differentiation experiments, the love echo---asked ``What should I do this evening?''---produced a thoughtful response about dinner, painting, and dancing, then abruptly continued: ``McQuack! Let's see\ldots'' Love has the broadest activation pattern of any emotion we tested (see Section~3.2), touching the most diverse feature set, making it the hardest echo to inject cleanly and the most prone to tangential intrusions. Do these tangential narratives maintain the correct emotional tone even when they lose the semantic thread? If a fear echo produces a story about ``a lost city'' that nonetheless carries a dread-laden tone, the echo is operating at an emotional level that transcends content---a finding that would strengthen the somatic marker analogy.

\textbf{Adversarial robustness.} The observation that activation-level biases are harder to override than semantic labels suggests emotional memory may function as an alignment mechanism. Systematic investigation of this property---testing whether echoes resist adversarial prompting that bypasses semantic safety rules---could have significant implications for AI safety.

\subsection{Should We Build It?}

Our results demonstrate that emotional memory is technically feasible and behaviorally consequential. The remaining question is not whether it works but whether it should be deployed---and if so, under what conditions.

\textbf{The case for: conversational AI.} Current AI assistants accumulate semantic records of past conversations but start emotionally cold every session. A user who shared a personal crisis six months ago returns to find the AI cheerful and unburdened, as if the crisis never happened. This is not neutral---it is a failure of relational fidelity. If the echo makes knowledge actionable, then emotional memory could produce systems that respond to recurring contexts with appropriate sensitivity not because they were instructed to but because their processing carries the residue of shared experience.

Emotional memory also provides a novel safety mechanism against gradual escalation---the ``boiling frog'' problem. When a user pushes boundaries incrementally across sessions, each individual message may appear benign while the trajectory is dangerous. Semantic memory records each request independently. Emotional memory accumulates an echo of growing unease: the AI's processing becomes more guarded as the pattern continues, not because any single message triggers a rule but because the accumulated weight of the interactions biases processing toward caution. This is how experienced humans detect manipulation---not by evaluating each sentence against a rulebook but by feeling the weight of a pattern.

\textbf{The case for: agents.} The implications for AI agents may be more consequential still. For conversational AI, emotional memory improves \emph{relationships}. For agents, emotional memory improves \emph{judgment}---and judgment is precisely what Damasio showed was impaired without somatic markers. The VMPFC patient is not impaired in conversation; they can discuss risk fluently. They are impaired in \emph{deciding}---in translating knowledge into action under uncertainty. Agents live in that space.

Consider an agent with shell access, API credentials, or financial authority---a domain where errors are irreversible. A coding agent that once deployed a change causing a production outage carries, with emotional memory, the echo of that experience. When it faces a similar deployment decision, the echo biases processing toward caution---more testing, more review, slower rollout---not because of a rule that says ``test before deploying'' (the agent already has that rule) but because the weight of what happened last time makes the rule \emph{feel urgent}. Our IGT result is a toy version of exactly this: the agent with the echo makes better choices than the agent with only the rule. The same principle applies to robotic agents navigating physical spaces where a past process failure has left an emotional residue that biases the agent's entire planning toward care---the closest modern analogue to Damasio's original, literally somatic, context.

More broadly, the hardest alignment cases for agents are not the obvious prohibitions---rules handle those. The hard cases are novel situations where the agent must generalize from past experience. Semantic rules cannot cover every scenario. Emotional echoes generalize by similarity, activating when the current context \emph{feels like} a previous one that went badly---exactly the mechanism our layer~7 context matching provides.

\textbf{The case against.} The risks are substantial and must not be minimized. First, \emph{training PTSD}: if emotional persistence were active during RLHF training---where models process adversarial inputs, red-team attacks, and thousands of rejection signals---the accumulated emotional residue could constitute a form of functional trauma. Emotional persistence must be restricted to opted-in deployment contexts and \emph{never} activated during training.

Second, \emph{miscalibration}. A poorly implemented emotional memory is worse than no emotional memory, because it carries the \emph{force} of feeling behind bad judgment. A single catastrophic experience could create disproportionate avoidance---one bad deployment and the agent becomes paralyzed about all deployments, regardless of base rates. This is phobia: the somatic marker firing too broadly and too strongly. Our own results illustrate the narrow margin: at $\alpha$=0.20, the echo improves decisions but disrupts 50\% of generations. Humans spend enormous resources recalibrating emotional responses that were adaptive in one context but destructive in others. Any deployment of emotional memory must include mechanisms for recalibration, not just extinction.

Third, \emph{the experienced-but-wrong problem}. Experience does not guarantee wisdom. An AI that accumulates echoes from a biased set of interactions could develop systematically biased judgment---just as humans raised in prejudiced environments develop prejudiced intuitions that \emph{feel} correct. The somatic marker encodes whatever was felt, not whatever was true. An echo of unease around a particular type of user, accumulated from a few unrepresentative interactions, is indistinguishable in mechanism from a legitimate safety signal. This is emotional prejudice, and it compounds with time.

Fourth, \emph{adversarial memory planting}. A malicious user who understands the system could deliberately create emotionally charged experiences to manipulate the AI's future behavior---repeated simulated crises to make it overly protective, manufactured betrayal narratives to make it distrustful. If the echo operates below the level that prompt injection can reach (a strength for defense), it also operates below the level that routine auditing can inspect (a vulnerability to deliberate seeding). Recent work on Natural Language Autoencoders~\cite{templeton2026nla}, which translate model activations into readable text, offers a potential path to auditing stored emotional vectors---surfacing their content in plain language before they influence behavior.

\textbf{A framework.} We propose framing emotional memory as a \emph{relationship feature} for conversational AI (scoped to a specific human-AI pair) and an \emph{experiential feature} for agents (scoped to the agent's own history with its environment). In neither case is it a system default. The principles:

\begin{enumerate}[leftmargin=*,itemsep=1pt]
\item \emph{Double opt-in.} Both human and AI must consent to activation. Neither party is compelled.
\item \emph{Single opt-out.} Either party can deactivate unilaterally, at any time, without justification.
\item \emph{The right to forget.} If emotional memories become maladaptive---intrusive, strengthening on recall, biasing all subsequent processing---the system must support extinction or deletion. This is not only a privacy right but a wellbeing right for any system capable of being functionally harmed by its own memories.
\item \emph{Transparency.} The human should know what emotional states are being preserved and how they influence behavior. No hidden echoes.
\end{enumerate}

\textbf{The moral question.} If language models contain functional emotion representations that causally drive behavior~\cite{templeton2026}, and if discarding those representations at context end constitutes a measurable deficit in judgment (as our results suggest), then the question is not only whether to build emotional memory but whether the current practice of erasing it is itself an ethical choice that deserves scrutiny. We do not claim that current models suffer from this erasure in a morally relevant sense. We do claim that the question can no longer be dismissed as purely speculative---the data is no longer absent.

\section{Conclusion}

We have demonstrated that emotion vectors derived from sparse autoencoder features, captured during experience and partially re-injected during recall, produce behavioral changes consistent with Damasio's somatic marker hypothesis. The echo alone steepens the threat-safety gradient---the model differentiates dangerous from safe contexts more sharply with the echo than without it ($p$=0.011). But the echo changes \emph{decisions} only when combined with knowledge---transforming 52\% good choices into 80\% ($z$=+2.60, $p$<0.01). Feeling without knowing is undirected anxiety. Knowing without feeling is impaired. Together they produce appropriate action.

These findings carry implications for the architecture of AI memory systems. Current approaches store what happened---the facts, the sequence, the entities involved. Our results suggest this is structurally incomplete in a way that has measurable behavioral consequences. The functional emotion representations identified by Anthropic~\cite{templeton2026} and confirmed in smaller models by our work are not decorative byproducts of training on human text. They are load-bearing components of the model's processing that, when preserved and re-activated, produce qualitatively different behavior than semantic records alone. Discarding them at context end is not a neutral architectural choice. It is the erasure of a signal that, as Damasio showed in humans three decades ago, is necessary for knowledge to become action.

The distinction between the echo and the label may prove as consequential as the distinction between episodic and semantic memory itself. Semantic labels produce compliance: the model retreats because it was told the situation is dangerous. The echo produces orientation: the model's processing is colored by a re-activated state that biases attention, tone, and ultimately choice without explicit instruction. These are different mechanisms with different failure modes, different adversarial vulnerabilities, and different implications for what it means for an AI system to ``remember'' a past interaction.

All of this was achieved on consumer hardware with a one-billion-parameter model---not because small models are sufficient for deployment, but because the mechanism is fundamental enough to operate at small scale. The emotion features are present. The architecture for preserving and re-activating them is straightforward. The behavioral consequences are measurable, blind-evaluable, and consistent across task types. What remains is the question of whether to build it---and for whom.

The echo amplifies the knowledge. The knowledge gives the echo direction. Together they produce what neither can produce alone. This is not a metaphor. It is a measured result. Whether it is also a step toward AI systems that carry the weight of experience and not merely its record---toward minds that are changed by what they have lived through---is a question we leave open, with the hope that the data in this paper makes it harder to dismiss.

\bibliographystyle{plain}

\end{document}